\documentclass{article}

\usepackage[utf8]{inputenc} % allow utf-8 input
\usepackage[T1]{fontenc}    % use 8-bit T1 fonts
\usepackage{hyperref}       % hyperlinks
\usepackage{url}            % simple URL typesetting
\usepackage{booktabs}       % professional-quality tables
\usepackage{amsfonts}       % blackboard math symbols
\usepackage{nicefrac}       % compact symbols for 1/2, etc.
\usepackage{microtype}      % microtypography
\usepackage{comment}        % comment

\usepackage{algorithm}
\usepackage{algorithmic}
\usepackage{enumitem}
\usepackage{graphicx}
\usepackage{amssymb}
\usepackage{xcolor}
\usepackage{dsfont} % for identity operator

\usepackage{pifont}% http://ctan.org/pkg/pifont
\usepackage{caption}
\usepackage[tableposition=above]{caption} % for the caption to be placed nicely below.
\usepackage{subcaption}

\usepackage{comment}

\usepackage{amsmath,amsfonts,bm}

\def\eqref#1{equation~\ref{#1}}
\def\1{\bm{1}}

\DeclareMathAlphabet{\mathsfit}{\encodingdefault}{\sfdefault}{m}{sl}
\SetMathAlphabet{\mathsfit}{bold}{\encodingdefault}{\sfdefault}{bx}{n}

\newcommand{\E}{\mathbb{E}}

\DeclareMathOperator{\id}{\mathds{1}}

\usepackage[accepted]{icml2020}

\icmltitlerunning{Practical Bayesian Neural Networks via Adaptive Optimization Methods}
\begin{document}

\twocolumn[
\icmltitle{Practical Bayesian Neural Networks via Adaptive Optimization Methods}



\icmlsetsymbol{equal}{*}

\begin{icmlauthorlist}
\icmlauthor{Samuel Kessler}{equal,ox}
\icmlauthor{Arnold Salas}{equal,ox}
\icmlauthor{Vincent W.~C.~Tan}{ox}
\icmlauthor{Stefan Zohren}{ox}
\icmlauthor{Stephen J.~Roberts}{ox}
\end{icmlauthorlist}

\icmlaffiliation{ox}{Oxford University}

\icmlcorrespondingauthor{Samuel Kessler}{skessler@robots.ox.ac.uk}

\icmlkeywords{Machine Learning, ICML}

\vskip 0.3in
]



\printAffiliationsAndNotice{}  % leave blank if no need to mention equal contribution

\begin{abstract}
We introduce a novel framework for the estimation of the posterior distribution over the weights of a neural network, based on a new probabilistic interpretation of adaptive optimisation algorithms such as \textsc{AdaGrad} and \textsc{Adam}. We demonstrate the effectiveness of our  Bayesian \textsc{Adam} method, \textsc{Badam}, by experimentally showing that the learnt uncertainties correctly relate to the weights' predictive capabilities by weight pruning. We also demonstrate the quality of the derived uncertainty measures by comparing the performance of \textsc{Badam} to standard methods in a Thompson sampling setting for multi-armed bandits, where good uncertainty measures are required for an agent to balance exploration and exploitation. 
\end{abstract}

\section{Introduction}
\label{sec:intro}

Exact Bayesian inference over the weights of a neural network is intractable as the number of parameters is very large and the functional form of a neural network does not lend itself to exact integration. For this reason, much of the research in this area has been focused on approximation techniques. Approximate inference methods that scale to large models include the Laplace approximation together with various different approximations to the Hessian of the log posterior. For instance a diagonal approximation is remarkably effective for Continual Learning (CL) \cite{Kirkpatrick}, the Kronecker factored Laplace approximation introduces covariances between weights of the same layer of a Neural Network (NN) \cite{Ritter2018}. Variational inference (VI) methods have also been able to scale \cite{graves11, 1505}. Recent advances in reparameterisation gradients \cite{Kingma_VAE, Figurnov} and automatic differentiation libraries have seen enormous advances in VI approximate inference techniques. However, VI methods require longer training times, have twice the number of parameters and are difficult to implement in comparison to Maximum a Posteriori (MAP) estimation of NNs. Natural gradient variational inference methods of \cite{zhang, vadam} help overcome the problem of slow convergence and difficulty in implementation by utilising the information geometry of the underlying parameter manifold which results in simple update equations. Variational Dropout \cite{Gal2016}, has found enormous success due to ease of implementation by interpreting a dropout regularised network as Gaussian Processes (GPs) on each layer of a NN. However, an explicit approximate posterior is marginalised automatically, and thus of limited use for CL, for instance.

\begin{figure*}[t]
     \centering
     \begin{subfigure}[b]{0.19\textwidth}
         \centering
         \includegraphics[width=\textwidth]{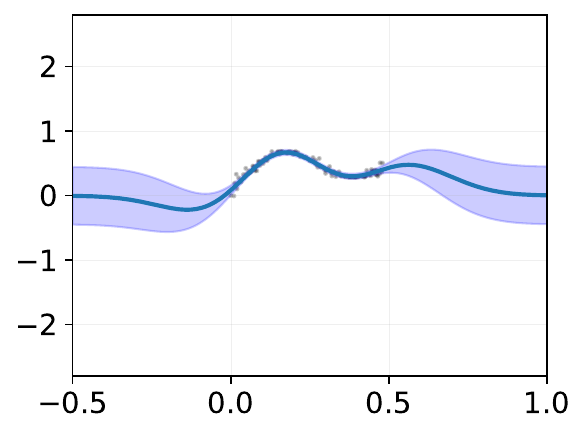}
         \caption{GP}
         \label{fig:re_gp}
     \end{subfigure}
     %\hfill
     \begin{subfigure}[b]{0.19\textwidth}
         \centering
         \includegraphics[width=\textwidth]{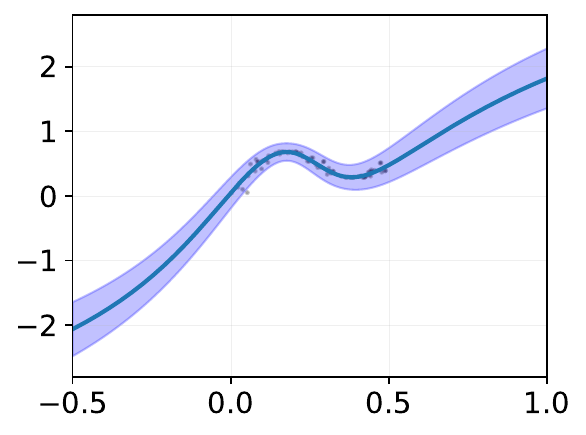}
         \caption{Badam}
         \label{fig:reg_badam}
     \end{subfigure}
     %\hfill
     \begin{subfigure}[b]{0.19\textwidth}
         \centering
         \includegraphics[width=\textwidth]{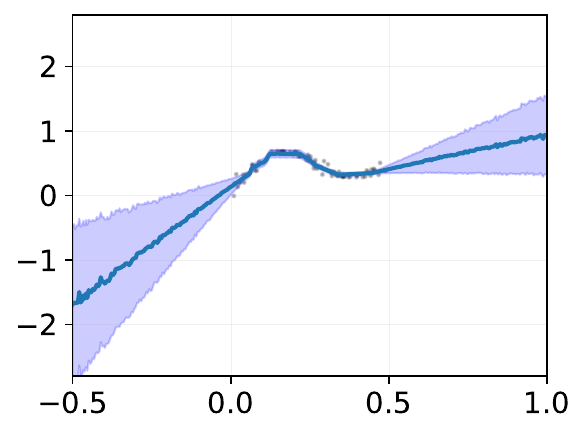}
         \caption{MC dropout}
         \label{fig:reg_mc_dropout}
     \end{subfigure}
     %\hfill
     \begin{subfigure}[b]{0.19\textwidth}
         \centering
         \includegraphics[width=\textwidth]{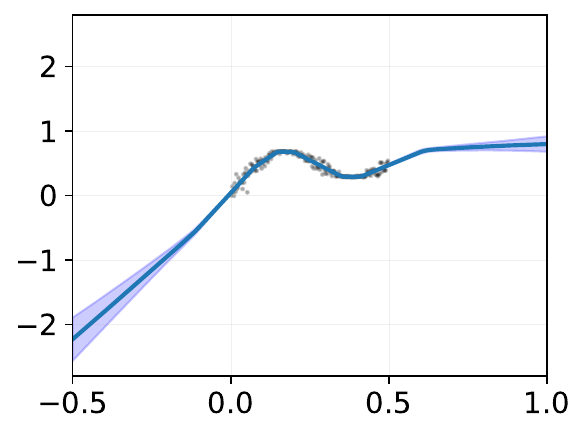}
         \caption{BBB}
         \label{fig:reg_bbb}
     \end{subfigure}
     %\hfill
     \begin{subfigure}[b]{0.19\textwidth}
         \centering
         \includegraphics[width=\textwidth]{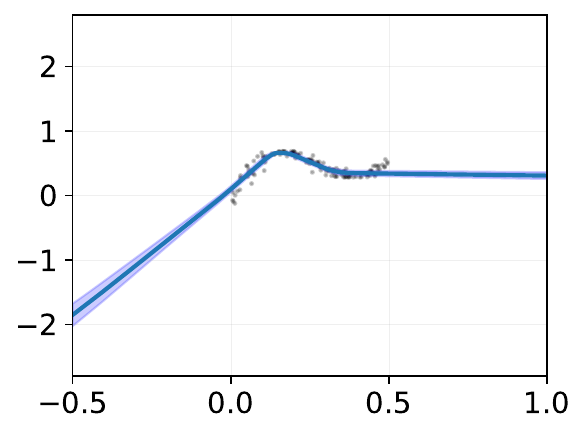}
         \caption{Vadam}
         \label{fig:reg_bbb}
     \end{subfigure}
        \caption{Qualitative comparison of the predictive uncertainties of our method and our principle baselines. \textsc{Badam}, like the Laplace approximation, places some probability mass where data has been observed and less further away \cite{Ritter2018}.}
        \label{fig:reg}
\end{figure*}

In this paper, we develop a novel Bayesian approach to learning for neural networks, built upon adaptive methods such as \textsc{AdaGrad} \cite{adagrad}, \textsc{RMSProp} \cite{rmsprop} and \textsc{Adam} \cite{adam}. Our method relies on a new probabilistic interpretation of adaptive algorithms, that effectively shows these can readily be utilised together with a Laplace approximation of the posterior. Our proposed algorithm is also similar in spirit to the work of \cite{vadam}, but in contrast, their algorithm performs natural gradient variational inference, implemented within \textsc{Adam} via weight perturbation of the gradient evaluation. Examples of the predictive uncertainties of our method and other common methods for approximate inference of Bayesian Neural Networks (BNNs) for a simple regression task are show in Figure~\ref{fig:reg} and discussed in Section~\ref{sec:results}.

\section{Preliminaries}

%\paragraph{Notation.} Vectors are denoted by lower case Roman letters such as $a$, and all vectors are assumed to be column vectors. Upper case roman letters, such as $M$, denote matrices, with the exception of the identity matrix which we denote by $\id$ and whose dimension is implicit from the context. Finally, for any vector $g_i\in\mathbb{R}^d$, $g_{i,j}$ denotes its $j$th coordinate, where $j\in[d]$.% and $[N] \equiv \{1, 2, \ldots, N\}$.

\paragraph{Problem setup.} Let $f(\theta)$ be a noisy objective function, a scalar function that is differentiable w.r.t. the parameters $\theta\in\Theta$, where $\Theta$ denotes the parameter space.% In general, $\Theta$ is a subset of $\mathbb{R}^d$, but for simplicity, we shall assume that $\Theta = \mathbb{R}^d$ throughout the remainder of this paper. 
We are interested in minimising the expected value of this function, $\mathbb{E}[f(\theta)]$, w.r.t. its parameters $\theta$. Let $f_1(\theta), \ldots, f_T(\theta)$ denote the realisations of the stochastic function at the subsequent time steps $t \in [T]$. The stochastic nature may arise from the evaluation of the function at random subsamples (minibatches) of datapoints, or from inherent function noise.

The simplest algorithm for this setting is the standard online gradient descent algorithm \cite{ogd}, which moves the current estimate $\theta_t$ of $\theta$ in the opposite direction of the last observed (sub)gradient value $g_t = \nabla f_t(\theta_t)$, i.e.,
\begin{equation}
	\theta_{t+1} = \theta_t - \eta_t g_t,
\end{equation}
where $\eta_t > 0$ is an adaptive learning rate that is typically set to $\eta/\sqrt{t}$, for some positive constant $\eta$. While the decreasing learning rate is required for convergence, such an aggressive decay typically translates into poor empirical performance. 

\paragraph{Generic adaptive optimisation methods.}

We now present a framework that contains a wide range of popular adaptive methods as special cases, and highlight their differences. The presentation here follows closely that of \cite{amsgrad}. The update rule of this generic class of adaptive methods can be compactly written in the form
\begin{equation}
\label{eq:sub_grad_descent}
	\theta_{t+1} = \theta_t - \eta_t V_t^{-1/2}m_t,
\end{equation}
where $m_t$ and $V_t$ are estimates of the (sub)gradient and inverse Hessian, respectively, of the functions $f_t(\cdot)$, based on observations up to and including iteration $t$. In other words, they are functions of the (sub)gradient history $g_{1:t} \equiv g_1, \ldots, g_t$, which we express as
\begin{align}
	m_t = \widehat{g}_t(g_{1:t}), \qquad
	V_t^{1/2} = \widehat{H}_t(g_{1:t}),
\end{align}
where $\widehat{g}_t(\cdot)$ and $\widehat{H}_t(\cdot)$ denote approximation functions for the (sub)gradient and Hessian of the loss function at iteration $t$, respectively. The corresponding procedure is repeated until convergence.

% \footnote{The Hessian of mean-squared error loss functions $f(\theta) = \frac{1}{2}\sum_{n}(\hat{y}_n(x_n, \theta) - y_n)^2$ can be approximated by a sum outer products of the Jacobians.}

% \scom{Vincent: The use of the inverse of the square root of the Hessian matrix as the preconditioner implies that the conditional variance of the gradient equals the unity matrix. It has the convenient effect that the variability of the gradient is determined solely by the learning rate.}

% \begin{comment}
% \begin{algorithm}
% \caption{Generic Adaptive Subgradient Descent}
% \label{alg:generic-adaptive-method-setup}
% \begin{algorithmic}
% \STATE{\bfseries Input:} $\theta_1 \in \mathbb{R}^d$, learning-rate schedule $\{\eta_t\}_{t=1}^T$, sequence of (sub)gradient and Hessian estimators $\{\widehat{g}_t(\cdot),\, \widehat{H}_t(\cdot)\}_{t=1}^T$
% \FOR{$t=1$ {\bfseries to} $T-1$}
% 	\STATE $g_t = \nabla f_t(\theta_t)$
% 	\STATE $m_t = \widehat{g}_t(g_{1:t}) \quad \text{and} \quad V_t^{1/2} = \widehat{H}_t(g_{1:t})$
% 	\STATE $\theta_{t+1} = \theta_t - \eta_t V_t^{-1/2} m_t$
% \ENDFOR
% \STATE{\bfseries Output:} $\theta_{T}$
% \end{algorithmic}
% \end{algorithm}
% \end{comment}

%For computational performance many popular algorithms restrict themselves to diagonal variants of adaptive subgradient descent, such that $V_t = \mathrm{diag}(v_t)$, where $v_t$ is the vector of diagonal elements. 
We first observe that the standard online gradient descent (\textsc{OGD}) algorithm arises as a special case of this framework if we use:
\begin{equation}
	m_t = g_t, \qquad V_t = \id.
\end{equation}
%Other variants use a decay of the learning rate, which corresponds to $\psi_t(g_1, \ldots, g_t) = t\boldsymbol{I}$ for all $t\in[T]$. While the decreasing step size is required for convergence, such an aggressive decay typically translates into poor empirical performance.
The key idea of adaptive methods is to choose estimator functions appropriately so as to entail good convergence. For instance, \textsc{AdaGrad} \cite{adagrad}, employs the following estimator functions:
\begin{equation}
	m_t = g_t, \qquad
	V_t = \frac{1}{t} \mathrm{ diag}\left(\sum_{i=1}^t g_i g_i^{\top}\right).
\end{equation}
In contrast to the learning rate of $\eta/\sqrt{t}$ in \textsc{OGD} with learning-rate decay, such a setting effectively implies a modest learning-rate decay of $\eta/\sqrt{\sum_i g_{i,j}^2}$ for $j\in[d]$. When the gradients are sparse, this can potentially lead to huge gains in terms of convergence (see \cite{adagrad}). % These gains have also been observed in practice even in some non-sparse settings.

\paragraph{Adaptive methods based on EWMA.}

Exponentially weighted moving average (EWMA) variants of \textsc{AdaGrad} are popular in the deep learning community. \textsc{AdaDelta} \cite{adadelta}, \textsc{RMSProp} \cite{rmsprop} and \textsc{Adam} \cite{adam} are some prominent algorithms that fall in this category. The key difference between these algorithms and \textsc{AdaGrad} is that they use an EWMA as the function $V_t$ instead of a simple average. \textsc{Adam}, a particularly popular variant, is based on the following estimator functions:
\begin{align}
\begin{split}
	m_t &= \frac{1-\beta_1}{1-\beta_1^t}\sum_{i=1}^t \beta_1^{t-i}g_i, \\
	V_t &= \frac{1-\beta_2}{1-\beta_2^t}\mathrm{diag}\left(\sum_{i=1}^t \beta_2^{t-i}g_i g_i^\top\right),
\end{split}
\end{align}
where $\beta_1, \beta_2 \in [0, 1)$ are exponential decay rates. % This update can alternatively be stated in terms of the following simple recursions:
% \begin{align}
% \begin{split}
% 	m_{t,i} &= \frac{\beta_1 m_{t-1,i} + (1-\beta_1)g_{t,i}}{1-\beta_1^t}, \\
% 	v_{t,i} &= \frac{\beta_2 v_{t-1,i} + (1-\beta_2)g_{t,i}^2}{1-\beta_2^t},
% \end{split}
% \end{align}
% for all $t\in[T]$, with $m_{0,i} = v_{0,i} = 0$ for all $i\in[d]$. Note that the denominator represents a bias-correction term. A value of $\beta_1=0.9$ and $\beta_2=0.999$ is typically recommended in practice \cite{adam}.
%We note the additional projection step in Algorithm~\ref{alg:generic-adaptive-method-setup} in comparison to \textsc{Adam}. When $\Theta=\mathbb{R}^d$, the projection operator is equivalent to the identity operator and this corresponds to the algorithm in \cite{adam}.
\textsc{RMSProp}, which appeared in an earlier unpublished work \cite{rmsprop}, is essentially a variant of \textsc{Adam} with $\beta_1=0$. In practice, especially in deep-learning applications, the momentum term arising due to non-zero $\beta_1$ appears to significantly boost performance.

% \begin{comment}
% More recently, \citet{amsgrad} pointed out that the aforementioned methods fail to converge to an optimal solution (or a critical point in non-convex settings). They showed that one cause for such failures is the use of EWMAs and, as a result of this, proposed \textsc{AMSGrad}, a variant of \textsc{Adam} which relies on a long-term memory of past gradients. Specifically, the \textsc{AMSGrad} update rule is characterised by the following system of recursive equations:
% \begin{equation}
% \tag{\textsc{AMSGrad}}
% \begin{split}
% 	m_{t} &= \beta_{1t} m_{t-1} + (1-\beta_{1t})g_{t} \\
% 	v_{t} &= \beta_2 v_{t-1} + (1-\beta_2)g_{t}g_{t}^{\top}
% 	\\
% 	\hat{v}_{t,i} &= \max(\hat{v}_{t-1,i}, \, v_{t,i}) 
% 	\\
% 	V_t &= \mathrm{diag}(\hat{v}_{t,1}, \ldots, \hat{v}_{t,d}),
% \end{split}
% \end{equation}
% for all $t\in[T]$, with $m_{0,i} = v_{0,i} = \hat{v}_{0,i} = 0$ for all $i\in[d]$, and where $\{\beta_{1t}\}_{t=1}^T$ is a sequence of exponential smoothing factors.
% \end{comment}

More recently, \cite{vadam} are able to derive similar EWMA update rules by performing natural gradient conjugate computations based on \cite{Khan2017} to Bayesian neural networks using a Gaussian approximation with a mean-field approximation for the Hessian along with squared gradients. By exploiting the duality within the exponential family distributions, simple updates for variational BNNs with fast convergence are derived.

\paragraph{Bayesian neural networks.}

A Bayesian neural network (BNN) is a neural network where the weights, $\theta$ are random variables. Consider an i.i.d.\ data set of $N$ feature vectors $x_1, \ldots, x_N\in\mathbb{R}^d$, with a corresponding set of outputs $\{y_1, \ldots, y_N\} \in \mathbb{R}$. For illustration purposes, we shall suppose that the likelihood for each datapoint is Gaussian, with mean $\mathrm{NN}(x, \theta)$ and variance $\sigma_o^2$:
\begin{equation}
	p(y_n|x_n, \theta, \sigma^2)
	= \mathcal{N}(y_n|\mathrm{NN}(x_n, \theta), \, \sigma_o^2).
\end{equation}
Similarly, we shall choose a prior distribution over the weights $\theta$ that is Gaussian of the form $p(\theta|\alpha) = \mathcal{N}(\theta|0, \, \sigma^2\id)$. The resulting posterior distribution is then
\begin{equation}
	p(\theta|\mathcal{D}, \sigma^2, \sigma_n^2)
	\propto p(\theta|\sigma^2)p(\mathcal{D}|\theta, \sigma_o^2),
\end{equation}
as a consequence of the nonlinear dependence of $\mathrm{NN}(x, \theta)$ on $\theta$, will be non-Gaussian. However, we can find a Gaussian approximation by using the Laplace approximation \cite{mackay92}.

\section{Probabilistic Interpretation of Adaptive Methods}

%We have touched upon the fact that adaptive subgradient methods use estimates of the curvature of the loss function for optimisation. Now we will explore how using these estimates one can obtain a posterior around the local minimum which the optimiser converges to, by a Laplace approximation of the loss. 
Consider a second-order Taylor expansion of the log-posterior $p(\theta| \mathcal{D})$ around the MAP $\theta_t$:
\begin{align}
\begin{split}
\label{eq:taylor}
	\log p(\theta | \mathcal{D})
	&\approx \log p(\theta_t | \mathcal{D}) \, + \, \frac{1}{2}(\theta - \theta_t)^{\top} H (\theta - \theta_t).
\end{split}
\end{align}

The gradient upon convergence is zero and so the first order gradient in the Taylor expansion drops out. $H = \E\left[ \nabla^2 \log p(\theta | \mathcal{D})|_{\theta=\theta_t} \right]$ and 
\begin{align}
\label{eq:posterior}
\nabla^2 \log p(\theta | \mathcal{D})|_{\theta=\theta_t}  & = \nabla^2 \log p(\mathcal{D} | \theta) |_{\theta=\theta_t}  + \sigma^2 \id
\end{align}
where the first term on the r.h.s is a Hessian of a mean-squared error loss function, $\log p(\mathcal{D} | \theta) \propto \frac{1}{2}\sum_{n}(\mathrm{NN}(x_n, \theta) - y_n)^2$, where $\sigma_o^2$ is treated a nuisance parameter and set to $1$. This Hessian can be approximated by the Gaussian-Newton matrix (GGN) \cite{mackay92}, (this approximation is introduced more formally in Section~\ref{sec:ggn}). The second term in Equation~(\ref{eq:posterior}) is the contribution from the Gaussian prior. The likelihood term is scaled by $N$ = $|\mathcal{D}|$ as most implementations will take an average of the log-likelihood. In practice, the GGN matrix requires iterating over the training data and calculating  and manipulating Jacobians which are expensive for large datasets and overparameterized NNs, notice that one can replace it by the approximation from adaptive optimisation methods which can also be used to train the NN:
\begin{equation}
	\E\left[ \nabla^2 \log p(\theta | \mathcal{D})|_{\theta=\theta_t} \right] \approx V_t ^{1/2}.
\end{equation}
Exponentiation of Equation~(\ref{eq:taylor}) yields a Gaussian functional form in $\theta$, and hence a Gaussian approximate posterior or Laplace approximation centred on the MAP:
\begin{equation}
\label{eq:finalposterior}
	q(\theta|\mathcal{D}, g_{1:t}, \eta_t, \omega)
	= \mathcal{N}\left( \theta \,|\,\theta_{t}, \,  \left (N V_t^{1/2} \right )^{-1} \right),
\end{equation}
where $\left (N V_t^{1/2} \right )^{-1}$ is guaranteed to be positive semi-definite, $\omega$ is a vector of hyperparameters\footnote{For example, $\omega = \emptyset$ in the case of \textsc{AdaGrad}, whereas $\omega = (\beta_1, \beta_2)$ for \textsc{Adam}.} other than the learning rate, if any, that govern the underlying adaptive method. We summarise the practical implementation of this procedure through our proposed algorithm, \textsc{Badam} in Section~\ref{sec:algo}. 

An advantage from this proposal is that $ V_t^{1/2}$ is readily available from the adaptive method, which offers computational and implementation benefits. In particular, the use of \textsc{Adam}'s EWMA estimates for $m_t$ and $V_t$, which enable us to filter out the noise arising from the stochastic nature of the gradients, while at the same time accounting for changes in these quantities over different areas of the loss landscape.
In contrast, using \textsc{Adagrad}'s estimate for $V_t$ would imply that the resulting covariance matrix in Equation~(\ref{eq:finalposterior}) is constant over the trajectory of the loss landscape: a rather severe assumption. In practice, we can run \textsc{Adam} and then get $V_t = \textrm{diag}(v_t)$ post-hoc to get a cheap approximate posterior. By re-using the curvature estimates we circumvent the need to calculate an approximation of the Hessian such as the GGN approximation which involves iterating over the training dataset and evaluating Jacobians. %The use of \textsc{Adam} will ensure a sensible posterior mean since it is equivalent to the point estimate generated by \textsc{Adam}. 

The estimate of the Hessian is a root mean square of gradients: $V^{1/2}_t$, contrast this with the GGN approximation of a typical Laplace approximation which does not contain the square root. The significance of the square root term is two-fold. Firstly, it ensures a reduction of the condition number for our estimate of the covariance in the high-dimensional setting. Thus, the diagonal estimate for the Hessian of the objective function has smaller eigenvalues, ensuring a more stable covariance upon inversion. Secondly, it ensures the Gaussian approximate posterior, $q(\theta|\mathcal{D}, g_{1:t}, \eta_t, \omega)$ has a more conservative covariance matrix and uncertainty estimate than the traditional Laplace which can place probability mass where the true posterior has none \cite{Ritter2018}.

%Finally, it is known that the Hessian of a neural network is ill-conditioned in the high-dimensional setting where the number of parameters is much larger than the number of data points. This results in the Laplace approximation overestimating the uncertainty severely underfitting. This effect on the predictive distribution is developed in Section~\ref{sec:pred_dist}. Hence, the estimate from adaptive sub-gradient methods will help to mitigate against the potential underfitting. %Additionally, it has been shown that by treating gradient descent as a linear dynamical system with noisy gradient observations and optimal weight values as a latent variable one can recover the root mean square normaliser update rule in adaptive subgradient methods \cite{Aitchison}.

\begin{figure}
  \centering
  \includegraphics[scale=0.3]{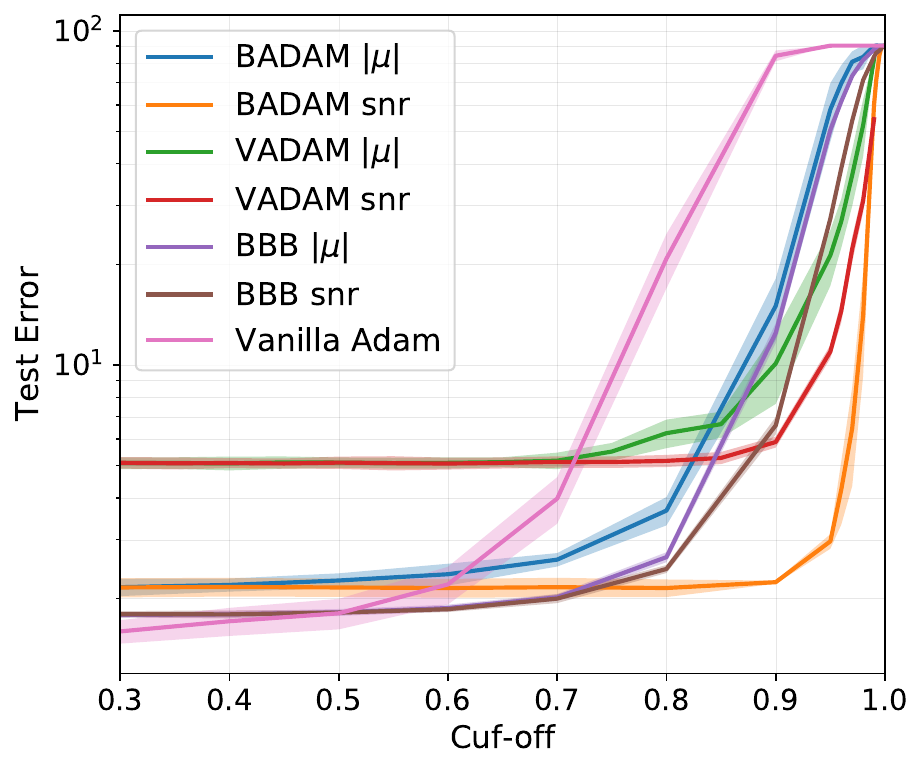}
  \caption{Weight pruning test accuracies. \textsc{Badam} produces high quality uncertainties and obtains high accuracies.}
  \label{fig:weight_pruning}
\end{figure}

\begin{table*}
  \centering
  \caption{Cumulative rewards normalised by the optimal reward $\pm$ one std. The results are taken from 10 runs. \textsc{Vadam} outperforms on most datasets, then \textsc{Badam} performs better than the other approximate inference methods.}
  \label{tab:bandits}
\begin{tabular}{llllllll}
\toprule
 \quad  & Mushroom & Financial & Jester & Statlog & Adult & Covertype & Census \\ % Wheel &
 \midrule
 \textsc{Badam} & $\bm{0.74}$\scalebox{0.75}{$\bm{\pm 0.09}$} & $0.74$\scalebox{0.75}{$\pm 0.03$} & $0.48$\scalebox{0.75}{$\pm 0.01$} &  $\bm{0.97}$\scalebox{0.75}{$\bm{\pm 0.02}$} & $0.13$\scalebox{0.75}{$\pm 0.01$} & $0.58$\scalebox{0.75}{$\pm 0.01$} & $0.36$\scalebox{0.75}{$\pm 0.06$} \\ % $0.71 \pm 0.09$ &
 \textsc{MC dropout} & $0.68$\scalebox{0.75}{$\pm 0.11$} & $0.68$\scalebox{0.75}{$\pm 0.02$} & $0.48$\scalebox{0.75}{$\pm 0.01$} & $0.96$\scalebox{0.75}{$\pm 0.02$} & $0.13$\scalebox{0.75}{$\pm 0.01$} & $0.57$\scalebox{0.75}{$\pm 0.01$} & $0.44$\scalebox{0.75}{$\pm 0.03$} \\ % $0.44 \pm 0.29$ &
 \textsc{BBB} & $0.60$\scalebox{0.75}{$\pm0.15$} & $0.21$\scalebox{0.75}{$\pm 0.13$} &  $0.46$\scalebox{0.75}{$\pm 0.02$} &   $0.71$\scalebox{0.75}{$\pm 0.17$} &  $0.10$\scalebox{0.75}{$\pm 0.04$} & $0.39$\scalebox{0.75}{$\pm 0.06$} & $0.38$\scalebox{0.75}{$\pm 0.02$}\\ % $0.51 \pm 0.26$ &
 \textsc{Vadam} & $\bm{0.76}$\scalebox{0.75}{$\bm{\pm0.03}$} & $0.81$\scalebox{0.75}{$\pm 0.00$} &  $\bm{0.48}$\scalebox{0.75}{$\bm{\pm 0.01}$} &   $\bm{0.99}$\scalebox{0.75}{$\bm{\pm 0.00}$} &  $\bm{0.26}$\scalebox{0.75}{$\bm{\pm 0.02}$} & $\bm{0.74}$\scalebox{0.75}{$\bm{\pm 0.01}$} & $\bm{0.65}$\scalebox{0.75}{$\bm{\pm 0.00}$}\\ % $0.51 \pm 0.26$ &
 Greedy & $0.64$\scalebox{0.75}{$\pm 0.18$} & $\bm{0.86}$\scalebox{0.75}{$\bm{\pm 0.03}$} &  $\bm{0.50}$\scalebox{0.75}{$\bm{\pm 0.01}$} & $\bm{0.97}$\scalebox{0.75}{$\bm{\pm 0.03}$} & $0.16$\scalebox{0.75}{$\pm 0.01$} & $0.62$\scalebox{0.75}{$\pm 0.03$} & $0.44$\scalebox{0.75}{$\pm 0.02$}\\ % $0.64 \pm 0.24$ &
Uniform & $-0.93$\scalebox{0.75}{$\pm 0.09$} &  $-0.01$\scalebox{0.75}{$\pm 0.01$} & $0.34$\scalebox{0.75}{$\pm 0.00$} & $0.14$\scalebox{0.75}{$\pm 0.00$} & $0.07$\scalebox{0.75}{$\pm 0.00$} & $0.14$\scalebox{0.75}{$\pm 0.00$} & $0.11$\scalebox{0.75}{$\pm 0.00$} \\ %& $0.28 \pm 0.00$
 \bottomrule
\end{tabular}
\end{table*}

%The term $V_t^{1/2}$ is used as an approximation for the curvature of the energy landscape. Intuitively, if the curvature is very high in a given direction then this leads to less variability in parameter space in this direction, on the other hand if the energy landscape is very flat one expects larger variability. The notion of curvature can thus be readily interpreted as a geometric notion of uncertainty. The mean is around the point estimate of our weight and hence and good posterior mean is ensured according to the final local minimum from \textsc{Adam}.

\section{Results}
\label{sec:results}

We demonstrate the approximate posterior's effectiveness by pruning on MNIST and in a contextual bandit setting. We present our method and compare to \textsc{BBB}, \textsc{MC dropout} and \textsc{Vadam}\footnote{Code is available at \url{https://github.com/skezle/BADAM}.}.

In Figure~\ref{fig:reg} we introduced our method's predictive distribution for a simple regression experiment. The predictive distribution for \textsc{Badam} is similar to a standard Laplace approximation in that more probability mass is placed is places close to the observations than the other methods and uncertainties widen gradually \cite{Ritter2018}. We use the Linearised Laplace trick for a stable predictive distribution \cite{Foong2019}. \textsc{BBB} and \textsc{Vadam} are expensive to run and at convergence don't produce large uncertainties far away from observations. \textsc{Vadam}, also requires a careful choice of an initial precision of the posterior to initialise the algorithm and its performance is sensitive to different specifications of the initial precision, which makes it difficult to train in practice.

% The code to reproduce these experiments is available at \url{github.com/skezle/BADAM}.

\paragraph{Classification on MNIST.} To assess the quality of the obtained uncertainties we follow weight-pruning on MNIST. Given a posterior mean $\mu$ and posterior variance $\sigma^2$, we sort the weights by their signal-to-noise ($|\mu| / \sigma$) ratio and discard the fraction $p$ of weights with the lowest values, by setting these weights to zero. As a baseline, we perform pruning on a model with $\Sigma = \id$. \textsc{Badam} can produce high quality uncertainties seen as the pruning via signal-to-noise drop in the accuracy is more robust to pruning than all other baselines and it has a high accuracy for without pruning. Experimental details can be found in Section~\ref{app:mnist}.

\paragraph{Contextual Bandits.} We demonstrate the effectiveness of the \textsc{Badam} uncertainties by using the BNN in a contextual multi-armed bandit setting where the agent requires a good measure uncertainty for decision making. The contextual multi-armed bandit problem proceeds as follows, at time $t$ our agent will receive a context $X_t \in \mathbb{R}^d$ and will need to decide which action $a_t \in \mathcal{A}$ to pick to maximise a total reward $r = \sum^{T}_{t=1} r_t$. Our agent learns a function $f: (X_t, a_t) \rightarrow r_t \in \mathbb{R}$. Thompson sampling provides a Bayesian framework for our agent to manage the exploration exploitation dilemma \cite{Thompson}. In Thompson sampling at each iteration our model samples from its prior and then greedily selects the action which maximises the reward. The model then receives feedback for the selected decision and updates its prior. No feedback is observed for actions which are not selected. The challenge in this setting is that at time $t$ our agent's approximate prior is used to estimate $r_t$ and pick an arm. The rewards $r_j$, $j \leq t$ are not i.i.d. and this feedback loop together with an approximate posterior can lead to disagreements between the model's posterior and the true posterior \cite{Riquelme}. This in turn can result in large cumulative regrets. Since the posterior uncertainty construction and representation learning in the \textsc{Badam} algorithm are separate we postulate good performance \cite{Riquelme}. Details of the experimental setup can be found in Section~\ref{app:bandits}.

The results in Table~\ref{tab:bandits} show that the rewards from \textsc{Badam} are comparable to \textsc{MC Dropout} and outperforms \textsc{BBB}. \textsc{Vadam} performs best, at each iteration of \textsc{Vadam} as there is an injection of noise which could increase exploration and explain the good performance. The greedy bandit performs well as in \cite{Riquelme}, however it underperforms \textsc{Badam} on the Mushroom bandit.

\section{Conclusion}
\label{sec:conclusion}
In this paper, we introduce a straightforward and cheap approach to posterior inference in BNNs, derived from a new probabilistic interpretation of adaptive optimisation methods. In particular, we discuss how to refine this framework to \textsc{Adam}. Finally, we demonstrate empirically the performance of the posterior distribution on pruning with MNIST classification and on Thompson sampling in contextual bandits. \textsc{Badam} is computationally efficient, like \textsc{MC dropout}, and does not require post-hoc computation of a covariance matrix like the Laplace approximation. It has the additional advantage of producing an explicit approximate posterior distribution like mean-field variational inference methods \cite{1505, vadam}, without the complexity of implementing and tuning the stochastic variational inference algorithms.

Since this work appeared, another post-hoc approach to posterior inference in NNs, Stochastic Weight Averaging Gaussian (SWAG) appeared \cite{Maddox2019}. SWAG approximates the posterior of the NN with a Gaussian whose moments are calculated by averaging the SGD iterates after convergence.

% Finally, we want to emphasise that the key underlying principle of the Bayesian treatment we propose in this paper is to provide a measure of uncertainty over a neural network's weight parameters, and not just a better or faster (in convergence terms) point estimate thereof. The quality of this uncertainty metric can be assessed by pruning the weights: the more robust the classification error of a Bayesian algorithm for learning neural networks is to weight pruning, the better the quality of the uncertainty embedded in the corresponding (approximate) weight posterior distribution will be. An algorithm achieving a smaller error at higher pruning rates (relative to pruning by $\mu$ only)-- even if its corresponding error rate at 0\% pruning is less attractive -- comes with genuinely desirable uncertainty estimates. This is clearly illustrated in Figure~\ref{fig:weight_pruning}. The quality of the uncertainty can also be judged on their performance on a Thompson sampling problem where decisions are made according the approximate posterior of our algorithms. That \textsc{Badam} is able to achieve similar performance to standard benchmarks is testament to the quality of its approximate posterior distribution.

\bibliography{example_paper}
\bibliographystyle{icml2020}

\newpage

\appendix
\onecolumn

\section*{Appendices}

\section{\textsc{Badam} Algorithm}
\label{sec:algo}

The specific approach whereby we can use \textsc{Adam} to obtain cheap estimates of the posterior are illustrated in Algorithm~\ref{alg:badam}.

\begin{algorithm}
\caption{\textsc{Badam}: Bayesian Learning of Neural Networks via \textsc{Adam}}
\label{alg:badam}
\begin{algorithmic}
\STATE{\bfseries Input:} $\theta_1 \in \mathbb{R}^d$, global learning rate $\eta$, exponential decay rates $\beta_1,\beta_2$, constant $\epsilon$
\STATE Set $m_{0} = v_0 = 0$
\FOR{$t=1$ {\bfseries to} $T-1$}
	\STATE $g_t = \nabla f_t(\theta_t)$
	\STATE $m_{t} = \frac{\beta_1 m_{t-1} + (1-\beta_1)g_{t}}{1-\beta_1^t}$ and $v_{t} = \frac{\beta_2 v_{t-1} + (1-\beta_2)g_{t}g_{t}^{\top}}{1-\beta_2^t}$
	\STATE $\theta_{t+1} = \theta_t - \eta m_t / (v_t^{1/2}+\epsilon)$ (element-wise division)
%	\STATE $\theta_{t+1} = \Pi_{\Theta, \, \sqrt{v_t}}(\hat{\theta}_{t+1})$ 
\ENDFOR
\STATE{\bfseries Output:} final weight distribution $\mathcal{N}\left( \theta \,|\, \theta_{t}, \, (N \mathrm{diag}(v_t^{1/2}))^{-1} \right)$
\end{algorithmic}
\end{algorithm}

\section{The Generalised Gauss Newton Matrix}
\label{sec:ggn}
The Hessian of the log-likelihood in Equation~(\ref{eq:posterior}) is the Hessian of a mean-squared error loss function, $\log p(\mathcal{D} | \theta) \propto \frac{1}{2}\sum_{n}(\mathrm{NN}(x_n, \theta) - y_n)^2$, ($\sigma_o^2$ is treated a nuisance parameter and set to $1$). The Hessian can be written as:
\begin{equation}
\begin{aligned}
    \nabla_{\theta}^{2}\log p(\mathcal{D}| \theta) &= \sum_{n} \nabla_{\theta} \mathrm{NN}(x_n, \theta) \nabla_{\theta} \mathrm{NN}(x_n, \theta)^{\top}
    + \sum_n (\mathrm{NN}(x_n, \theta) - y_n) \nabla^2_{\theta} \mathrm{NN}(x_n, \theta), 
\end{aligned}
\end{equation}
The second term involves a residual term which is assumed to be small for NNs if it is able to fit the data well. Hence the Hessian can be approximated by a sum of outer products of Jacobians of the NN predictions. Hence taking expectations like in Equation~(\ref{eq:posterior}):
\begin{align}
    \E_{p_{\theta}(y|x_n)} \left[ \nabla^2_{\theta} \log p(\mathcal{D}| \theta)\right] &= \E_{p_{\theta}(y|x_n)} \left[ \sum_{n} \nabla_{\theta} \mathrm{NN}(x_n, \theta) \nabla_{\theta} \mathrm{NN}(x_n, \theta)^{\top}  \right] \\
    &= \sum_{n} \nabla_{\theta} \mathrm{NN}(x_n, \theta) \nabla_{\theta} \mathrm{NN}(x_n, \theta)^{\top}, 
\end{align}
since the Jacobians do not depend on $y$ and then
\begin{align}
    H = \sum_{n} \nabla_{\theta} \mathrm{NN}(x_n, \theta) \nabla_{\theta} \mathrm{NN}(x_n, \theta)^{\top} + \sigma^2 \id.
\end{align}
This is referred to as the Gaussian-Newton matrix (GGN) \cite{mackay92}, its inverse is the covariance of traditional Laplace approximation of the posterior of a BNN. The GGN matrix corresponds Fisher information for exponential family distributions \cite{Martens2014}.

\section{Predictive Distributions}
\label{sec:pred_dist}
The predictive distribution over unseen data $x^*$ can be obtained through sampling from the posterior:
\begin{align}
p(y^*|x^*, \mathcal{D}) &= \int p(y^*|x^*, \theta) p(\theta|\mathcal{D}) d\theta \approx \frac{1}{M} \sum^{M}_{i=1}p(y^*| x^* \theta^{(i)}),
\end{align}
where $\theta^{(i)} \sim q(\theta | \mathcal{D})$. Sampling from the Laplace approximation can cause serious underfitting \cite{Lawrence2000}. Regularisation over a validation set can be performed to mitigate against underfitting by tuning $N$ and the prior variance $\sigma^2$. In practice the scaling $N$ can be treated as a hyperparameter \cite{Kirkpatrick}. Alternatively $N$ can be treated as ``pseudo-observations'' \cite{Ritter2018}. By adding dropout too neural network each new pass through the data will change the loss function landscape. With this in mind a heuristic it is useful to regularise the posterior by tuning $N$ according to a validation dataset. Alternatively the predictive distribution can be approximated by linearizing the output about the MAP to generate stable, $\theta_t$ \cite{Foong2019}. The predictive distribution for \textsc{Badam} uses this see Figure~\ref{fig:reg}.

\section{Experimental details}

\subsection{Toy Regression Experiments}
\label{app:toy_reg}

\paragraph{Data generation.} Data is generated from the function $y = x + 0.3 \sin(2\pi(x+\epsilon)) + 0.3 \sin(4\pi(x+\epsilon)) + \epsilon$ where $\epsilon \sim \mathcal{N}(0,0.02)$. Data for training is sampled from the range $(0.0, 0.5)$ while data used for evaluating the testing was sampled from the range $(-0.5, 1.2)$. The training set has $200$ samples.

\paragraph{Model architecture.} A two layer NN with ReLU activations and hidden state sixes of 50 neurons are used. For training \textsc{BADAM} we use and $L^2$ regularisation size of $10^{-4}$ and $N=200$ (in equation~\ref{eq:finalposterior}) is the number of points in the training set . the predictive distribution uses the Linearised-Laplace to obtain a closed form Gaussian solution to the predictive distribution \cite{Foong2019}. The \textsc{Badam} network is trained for $20,000$ epochs with a learning rate of $0.001$.

\subsection{MNIST Weight Pruning Experiments}
\label{app:mnist}
The \textsc{Badam} network $2$ layers of size $400$, uses dropout with a rate of $0.2$. For weight pruning it is the ordering of the weights which matters and so $N$ simply acts to scale the denominators of the snr, $|\mu| / \sigma$, therefore $N = 60,000$ which is the training set size. Interestingly setting no $L^2$ regularisation produced the best uncertainties. We found that using a larger value of $\beta_2=0.99999$ produces good uncertainties. We run \textsc{Badam} with default settings for $100$ epochs to achieve a good solution.

\subsection{Contextual Bandits Experiments}
\label{app:bandits}

\paragraph{Model step.} We use the experimental setup described in an implementation provided by \cite{Riquelme} for evaluation of our \textsc{Badam} algorithm. A detailed description of the datasets used for our multi-armed bandit experiments can also be found in the appendix of \cite{Riquelme}. We compare to \textsc{BBB}, \textsc{MC Dropout}, \textsc{Vadam} an NN which greedily picks each action (named Greedy in Tables~\ref{tab:bandits}), and as a baseline an agent which uniformly samples actions. The neural networks architectures used for all networks are the same: $2$ layers with $100$ units each and ReLU activations, they regress contexts in $\mathbb{R}^d$ to outputs in $\mathbb{R}^{|\mathcal{A}|}$. \textsc{Badam} weights are initialized with $U[-0.3, 0.3]$, and use gradient clipping such that the $\L2$-norm of the gradients are not greater than $5$. In terms of hyperparameters, all networks use an initial learning rate of $0.1$ with an inverse decaying schedule. The number of points seen $N$ is set dynamically as the $t\times \textrm{batch\_size}$, where $t$ is the number of rounds of the bandit.

\paragraph{Multi-armed bandits experimental details.} The specific implementation of \textsc{Badam} uses a Taylor approximation of the loss and with a Gaussian prior. Details are outlined in Section 3 of the previous version of this manuscript \url{https://arxiv.org/abs/1811.03679v2}. This was shown to work well in the contextual bandits setting rather than performing a Taylor expansion or Laplace around the MAP.

The experimental setup for the multi-armed bandits proceeds as follows: at each round a new context from the dataset is presented to the bandit algorithm, we go through the dataset once. Each action is initially selected $3$ times so that each agent has some initial information to learn from. Subsequently, actions are greedily chosen, and only the reward for the chosen action will be backpropagated upon training. In terms of training, all observed contexts, actions and rewards are stored in a buffer. The buffer is sampled to create batches used for training. Training occurred every $t_f=20$ rounds  for $t_s=50$ minibatches using a batch size of $512$ for all neural bandits.

The full datasets are used for all bandits experiments apart from the Census and Covertype datasets which are very large, we use a subset of $n=10000$ points from these two datasets.

\end{document}